# Model-Based Bayesian Reinforcement Learning in Large Structured Domains

**Stéphane Ross**
School of Computer Science
McGill University
Montreal, Canada

**Joelle Pineau**
School of Computer Science
McGill University
Montreal, Canada

## Abstract

Model-based Bayesian reinforcement learning has generated significant interest in the AI community as it provides an elegant solution to the optimal exploration-exploitation tradeoff in classical reinforcement learning. Unfortunately, the applicability of this type of approach has been limited to small domains due to the high complexity of reasoning about the joint posterior over model parameters. In this paper, we consider the use of factored representations combined with online planning techniques, to improve scalability of these methods. The main contribution of this paper is a Bayesian framework for learning the structure and parameters of a dynamical system, while also simultaneously planning a (near-)optimal sequence of actions.

## 1 Introduction

In the past decades, reinforcement learning (RL) has emerged as a useful technique for learning how to optimally control systems with unknown dynamics (Sutton & Barto, 1998). However classical RL has many shortcomings. In particular, RL does not address the problem of how to efficiently gather data to learn the parameters of the system, as well as how to behave in systems where the costs incurred during learning matter, i.e. the well known exploration-exploitation tradeoff problem. These shortcomings are mostly related to the fact that classical RL does not consider the uncertainty in the learned parameters for decision-making, nor does it allow for flexibly including prior knowledge about the system's dynamics.

Model-based Bayesian RL methods have successfully addressed these issues by maintaining a posterior distribution over unknown model parameters and acting such as to maximize long-term expected rewards with respect to this posterior (Dearden, Friedman, & Andre, 1999; Duff, 2002; Poupart, Vlassis, Hoey, & Regan, 2006). Prior knowledge of the system can be defined explicitly by specifying a prior distribution over model parameters. This allows for a flexible way of encoding uncertain knowledge into the learning algorithm. Furthermore, if the resulting decision problem is solved exactly, this provides an optimal exploration-exploitation tradeoff, in that the agent will behave such as to maximize long-term expected rewards with respect to the prior.

However, due to the high complexity of model-based Bayesian RL, most approaches have been limited to very small domains (10-20 states). This is mainly due to two reasons. First, when the number of states is large, a large amount of data needs to be collected to learn a good model, unless very few parameters are unknown or some structural assumptions are made to represent the dynamics with few parameters. Second, most planning approaches in Bayesian RL become intractable as the number of states increases, since planning is done over the full space of possible posteriors.

To address the first issue, we propose learning a factored representation of the dynamics via a Bayesian approach. Factored representations can efficiently represent the dynamics of a system with fewer parameters using a dynamic Bayesian network (DBN) that exploits conditional independence relations existing between state features (Boutilier, Dearden, & Goldszmidt, 2000; Guestrin, Koller, Parr, & Venkataraman, 2003). Bayesian RL techniques can be extended quite easily to factored representations when the structure of this DBN is known, however this is unreasonable in many domains. Fortunately, the problem of simultaneously learning the structure and parameters of a Bayes Net has received some attention (Heckerman, Geiger, & Chickering, 1995; Friedman & Koller, 2003; Eaton & Murphy, 2007), which we can leverage for our work. However while these approaches provide an effective way of learning the model, it is far from sufficient for Bayesian RL, where the goal is to *choose actions* in an optimal way, with respect to what we have learned about the model.

To address the issue of action selection, we propose incorporating an online Monte Carlo approach to evaluate sequences of actions with respect to the posterior over structures and parameters. The focus on online (rather than offline) planning means that we only need to plan with respect to the current posterior (rather than all possible posteriors), which offers substantial computational savings.

The main contribution of this paper is a novel Bayesian framework for optimizing the choice of actions in a structured dynamical system, with unknown structure and parameters. We present experimental results of our approach in a variety of large network administration domains, showing good performance for problems with thousands of states.

## 2 Background

A Markov Decision Process (MDP) is a general framework for decision making in stochastic systems (Bellman, 1957). It is often used to represent reinforcement learning problems (Sutton & Barto, 1998). We consider an unknown system represented by an MDP model in factored form $(S, A, T, R)$ where:

- $S : S_1 \times S_2 \times \cdots \times S_n$, is the (discrete) set of states of the system; $S_1, \ldots, S_n$ correspond to the domain of the $n$ state variables (features).
- $A$, the (discrete) set of actions that can be performed by the agent.
- $T : S \times A \times S \to [0, 1]$, the transition function, where $T(s, a, s') = \Pr(s'|s, a)$ represents the probability of moving to state $s'$ if the agent executes action $a$ in state $s$. This can be represented efficiently by a DBN for each action, exploiting conditional independence relations that exist between state features (Boutilier et al., 2000). For simplicity, we assume that these DBNs are bipartite graphs, so dependencies only exist between state variables at time $t$ and state variables at time $t+1$.
- $R : S \times A \to \mathbb{R}$, the reward function, defined for every action of the agent in every state.

The DBN defining $T$ for any action $a \in A$ is represented by a graph $G_a$ and set of parameters $\theta_{G_a}$ defining the conditional probability tables. For any state variable $s'_i$, we denote its set of parent variables in this graph by $\text{Par}_i(G_a)$ and given the previous state $s$, the values of these parents variables by $\text{ParVal}_i(s, G_a)$.

For each possible value $v \in S_i$ of state variable $s'_i$, and each possible assignment to its parent values $E \in S_{\text{Par}_i(G_a)} = \prod_{j \in \text{Par}_i(G_a)} S_j$, $\theta_{G_a}$ contains a parameter $\theta_{G_a}^{i,v|E}$ that defines $\Pr(s'_i = v|\text{ParVal}_i(s, G_a) = E, a)$. Given such graph $G_a$ and parameters $\theta_{G_a}$, $T(s, a, s')$ is computed efficiently as:

$$T(s, a, s') = \prod_{i=1}^{n} \Pr(s'_i|\text{ParVal}_i(s, G_a), a). \quad (1)$$

The goal of the MDP agent is to find an action selection strategy, called a *policy*, that maximizes its long-term expected rewards. The optimal action to take in a state $s$ is defined via the optimal value function $V^*$ representing the return obtained by the optimal policy starting in state $s$:

$$V^*(s) = \max_{a \in A} \left[ R(s, a) + \gamma \sum_{s' \in S} T(s, a, s') V^*(s') \right]. \quad (2)$$

The optimal action in $s$ is obtained by taking the arg max instead of the max in the last equation. In general, a factored representation of the transition does not induce a structured representation of the optimal value function. However, approximate algorithms exist to compute $V^*$ more efficiently by exploiting the factored representation (Guestrin et al., 2003).

### 2.1 Bayesian Reinforcement Learning

While the MDP framework allows one to compute the optimal policy for any stochastic system, it requires full knowledge of the transition dynamics. This is a strong assumption in practice. Model-based Bayesian RL weakens this assumption by instead maintaining a probability distribution over the possible settings of each unknown parameter (Dearden et al., 1999). It assumes an initial prior distribution over these parameters, and uses Bayes' rule to update the posterior distribution whenever state-transitions are observed in the course of interactions between the agent and the environment. Given that transition parameters are usually modeled using multinomial distributions, a natural choice to specify this posterior is the Dirichlet distribution. The Dirichlet is specified by "count" parameters, $\phi_1, \ldots, \phi_m$, specifying the likelihood $f(p|\phi)$ that outcomes $1, \ldots, m$ occur with probabilities $p_1, \ldots, p_m$ given they were observed $\phi_1, \ldots, \phi_m$ times:

$$f(p|\phi) = \frac{1}{B(\phi)} \prod_{i=1}^{m} p_i^{\phi_i - 1}, \quad (3)$$

where $B(\phi)$ is the multinomial beta function. This choice of prior allows for a flexible way to input prior knowledge in the system, as well as an easy way to maintain the posterior. We refer to the set of counts $\phi$ for all possible transitions $(s, a, s')$ as the information state of the agent. The resulting decision problem is the following: given the agent is in state $s$ with information state $\phi$, how should it behave such as to maximize its future expected rewards? This new decision problem can be modeled by an extended MDP model, called Bayes-Adaptive MDP (BAMDP), where the counts $\phi$ are included in the state space, and the transition function models how these parameters evolve given a particular state transition (Duff, 2002). This extended MDP has infinitely many states but can be solved exactly over a finite horizon for any particular current state and information state.

## 2.2 Learning Bayes Nets

Bayesian networks (BNs) have been used extensively to build compact predictive models of multivariate data. A BN models the joint distribution of multivariate data compactly by exploiting conditional independence relations between variables. It is defined by a set of variables $X$, a directed acyclic graph (DAG) structure $G$ over variables in $X$, and parameters $\theta_G$, where $\theta_G^{i,v|E}$ specifies the probability that $X_i = v$ given that its parents in $G$ take value $E$.

Several approaches exist to learn BNs. Learning a Bayes net can involve either only learning $\theta_G$ (if the structure $G$ is known), or simultaneously learning the structure $G$ and parameters $\theta_G$. For our purposes, we are mostly interested in Bayesian approaches that learn both the structure and parameters (Heckerman et al., 1995; Friedman & Koller, 2003; Eaton & Murphy, 2007). These Bayesian approaches proceed by first specifying a joint prior, $P(G, \theta_G)$, of the form:

$$P(G, \theta_G) = P(G)P(\theta_G|G), \quad (4)$$

where $P(G)$ is a prior over structures and $P(\theta_G|G)$ is a conditional prior on the parameters $\theta_G$ given a particular structure $G$. $P(G)$ is often chosen to be uniform, or proportional to $\beta^{|E(G)|}$ for some $\beta \in (0,1)$ where $|E(G)|$ is the number of edges in $G$, such as to favor simpler structures.

It follows that if dataset $D$ is observed, then the joint posterior is defined as follows:

$$P(G, \theta_G|D) = P(G|D)P(\theta_G|G, D). \quad (5)$$

To compute this posterior efficiently, several assumptions are usually made about the prior $P(\theta_G|G)$. First, it should factorize into a product of independent Dirichlet priors:

$$\begin{aligned} P(\theta_G|G) &= \prod_{i=1}^n \prod_{E \in S_{\text{Par}_i(G)}} P(\theta_G^{i,*|E}|G), \\ P(\theta_G^{i,*|E}|G) &\sim Dirichlet(\phi_G^{i,*|E}), \end{aligned} \quad (6)$$

Under this independence assumption, the term $P(\theta_G|G, D)$ is a product of Dirichlet distributions, which can be updated easily by incrementing counts $\phi_G^{i,v|E}$ for each $X_i = v|\text{ParVal}_i(G) = E$ in $D$.

A second common assumption is that two equivalent graph structures $G$ and $G'$ should have equivalent priors over $\theta_G$ and $\theta'_G$ (this is called the *likelihood equivalence assumption*). This enforces a strong relation between the priors $P(\theta_G|G)$ and $P(\theta_{G_c}|G_c)$ for the complete graph $G_c$ (where every variable depends on all previous variables). Hence specifying $\phi_{G_c}$ totally specifies the prior on $\theta_G$ for any other graph $G$.

For many problems, the posterior $P(G|D)$ cannot be maintained in closed form as it corresponds to a discrete distribution over $O(n!2^{\binom{n}{2}})$ possible graph structures. Instead, MCMC algorithms can be used to sample graph structures from this posterior (Friedman & Koller, 2003). The well known Metropolis-Hasting algorithm specifies that a move from graph $G$ to $G'$ should be accepted with probability $\min\left\{1, \frac{P(D|G')P(G')q(G|G')}{P(D|G)P(G)q(G'|G)}\right\}$, where $q(G'|G)$ is the probability that a move from $G$ to $G'$ is proposed and $P(D|G) = \int P(D|G, \theta_G)P(\theta_G|G)d\theta_G$. Such random walk in the space of DAGs has the desired stationary distribution $P(G|D)$. Under previous assumptions concerning the prior $P(\theta_G|G)$, $P(D|G)$ can be computed in closed form and corresponds to the likelihood-equivalence Bayesian Dirichlet score metric (BDe) (Heckerman et al., 1995). Typical moves considered include adding an edge, deleting an edge, or reversing an edge in $G$.

## 3 Bayesian RL in Factored MDPs

We consider the problem of acting optimally in a system represented as a factored MDP, in the case where both the structure and parameters of the DBNs defining the transition function, $T$, are unknown. We assume that the state features $S_1, \ldots, S_n$, the action set $A$, and the reward function $R$, are known. Our work extends trivially to the case where $R$ is unknown, but we leave this out for simplicity of presentation.

### 3.1 Factored Bayesian RL model

We consider the transition function $T$ as a hidden variable of the system, which is partially observed whenever state transitions occur in the system. In this view, the decision problem can be cast as a Partially Observable MDP (POMDP) (Kaelbling, Littman, & Cassandra, 1998). The state of this POMDP captures both the actual system state, and the DBNs defining $T$ for each action $a \in A$. Formally, this POMDP is defined

by the tuple $(S', A', Z', T', O', R')$:

- $S' : S \times \mathcal{G}^{|A|}$, where $S$ is the original state space of the MDP, $\mathcal{G}$ is the set of DBNs $(G, \theta_G)$ (one per action) and $G$ is a bipartite graph from $S_1, \ldots, S_n$ to $S_1, \ldots, S_n$.
- $A' = A$, the set of actions in the original MDP.
- $Z' = S$, the set of observations (i.e a transition to a particular state of the MDP)
- $T' : S' \times A' \times S' \to [0, 1]$, the transition function in this POMDP, where:

$$
\begin{aligned}
T'(s, G, \theta_G, a, s', G', \theta'_{G'}) & \quad (7) \\
= \quad & \Pr(s', G', \theta'_{G'} | s, G, \theta_G, a) \\
= \quad & \Pr(s' | s, G, \theta_G, a) \Pr(G', \theta'_{G'} | G, \theta_G, s, a, s').
\end{aligned}
$$

Since we assume that the transition function does not change over time, then

$$\Pr(G', \theta'_{G'} | G, \theta_G s, a, s') = I_{(G, \theta_G)}(G', \theta'_{G'})$$

(the indicator function of $(G, \theta_G)$), and

$$\Pr(s' | s, G, \theta_G, a) = \prod_{i=1}^{n} \theta_{G_a}^{i, s'_i | \text{ParVal}_i(s, G_a)}.$$

- $O' : S' \times A' \times Z' \to [0, 1]$, the observation function, where $O(s', G', \theta'_{G'}, a, z)$ is the probability of observing $z$ when moving to $(s', G', \theta'_{G'})$ by doing action $a$. Here we simply observe the state of the MDP, so $O(s', G', \theta'_{G'}, a, z) = I_{s'}(z)$.
- $R' : S' \times A' \to \mathbb{R}$, the reward function, which corresponds directly to the rewards obtained in the MDP, i.e. $R'(s, G, \theta_G, a) = R(s, a)$.

Given that the state is not directly observable (i.e. we do not know the correct structure and parameters), we maintain a probability distribution over states, called a *belief*. The initial belief state in this POMDP is the initial state of the environment, along with priors $P(G_a, \theta_{G_a}), \forall a \in A$. At time $t$, the belief state corresponds to the current state of the MDP, $s_t$, along with posteriors $P(G_a, \theta_{G_a} | h_t), \forall a \in A$, where $h_t$ is the history of actions and observations up to time $t$.

To represent this belief compactly, we assume that the joint priors $P(G_a, \theta_{G_a})$ satisfy the assumptions stated in section 2.2, namely they factorize into a product $P(G_a, \theta_{G_a}) = P(G_a) P(\theta_{G_a} | G_a)$ and the $P(\theta_{G_a} | G_a)$ are defined by a product of independent Dirichlet distributions. For each graph $G_a$, starting from prior counts $\phi_{G_a}^{i, v | E}$ for all state variables $i$, values $v \in S_i$, and parent values $E \in S_{\text{Par}_i(G_a)}$, the posterior counts are maintained by simply incrementing by 1 the counts $\phi_{G_a}^{i, s'_i | \text{ParVal}_i(s, G_a)}$ for all state variables $i$, each time a transition $(s, a, s')$ occurs. As mentioned in section 2.2, the main difficulty is in maintaining the posterior $P(G_a | h)$, which is infeasible when the space of graphs is large. We approximate this using a particle filter, and for each particle (i.e. a sampled graph $G_a$), the posterior $P(\theta_{G_a} | G_a)$ is maintained exactly with counts $\phi_{G_a}$. This particle filter is explained in more detail in the next section.

Finding the optimal policy for this POMDP yields an action selection strategy that optimally trades-off between exploration and exploitation such as to maximize long term expected return given the current model posterior and state of the agent. Our Bayesian RL approach therefore requires solving this POMDP. While many algorithms exist to solve POMDPs, few of them can handle high-dimensional infinite state spaces, as is required here. Hence, we propose to use online Monte Carlo methods to solve this challenging optimization problem (McAllester & Singh, 1999).

### 3.2 Online Monte Carlo Planning Algorithm

To solve the planning problem outlined above, we need efficient approximation methods, and in particular we turn to online sampling techniques to overcome the curse of dimensionality.

First, as mentioned above, we maintain the posterior $\Pr(G_a | h)$ using a particle filter algorithm. This is done by first sampling a set of $K$ graphs from the prior $P(G_a)$ for each action $a$. We assign each graph a probability, $p_a^j = \frac{1}{K}$, for $j = 1 : K$. For each sampled graph, we also have a product of Dirichlet priors on the parameters $\theta_{G_a}$. Whenever a transition $(s, a, s')$ occurs, the probability $p_a^j$ of graph $G_a^j$ is updated:

$$
\begin{aligned}
p_a'^{j} & = \frac{1}{\eta} p_a^j \int P(s' | s, a, G_a^j, \theta_{G_a^j}) P(\theta_{G_a^j} | G_a^j, h) d\theta_{G_a^j} \quad (8) \\
& = \frac{1}{\eta} p_a^j \prod_{i=1}^{n} \left[ \phi_{G_a^j}^{i, s'_i | \text{ParVal}_i(s, G_a^j)} / \sum_{v \in S_i} \phi_{G_a^j}^{i, v | \text{ParVal}_i(s, G_a^j)} \right]
\end{aligned}
$$

where the integral term is just the expected probability of $P(s' | s, a)$ under the current posterior for $\theta_{G_a^j}$, and $\eta$ is a normalization constant such that $\sum_{j=1}^{K} p_a'^{j} = 1$. For the Dirichlet posterior $P(\theta_{G_a^j} | G_a^j, h)$ associated with $G_a^j$, the appropriate counts are updated each time a corresponding state transition occurs.

Turning our attention to the planning problem, we now search for the best action to execute, given the current state, the current distribution on graphs (defined by $p_a^j$), and the current posterior over parameters for each graph. Define $Q^*(s, b, a)$ to be the maximum expected sum of rewards (i.e. the value) of applying action $a$ when the agent is in MDP state $s$ and has posterior $b$ over DBNs. Then the optimal value is defined by

$V^*(s, b) = \max_{a \in A} Q^*(s, b, a)$ and the best action to apply is simply $\arg\max_{a \in A} Q^*(s, b, a)$.

---

**Algorithm 1** $V(s, b, d, N)$

1: **if** $d = 0$ **then**
2:     **return** $\hat{V}(s, b)$
3: **end if**
4: $maxQ \leftarrow -\infty$
5: **for** $a \in A$ **do**
6:     $q \leftarrow R(s, a)$
7:     **for** $j = 1$ to $N$ **do**
8:         Sample $s'$ from $P(s'|s, b, a)$
9:         $b' \leftarrow \text{UPDATEGRAPHPOSTERIOR}(b, s, a, s')$
10:         $q \leftarrow q + \frac{\gamma}{N} V(s', b', d - 1, N)$
11:     **end for**
12:     **if** $q > maxQ$ **then**
13:         $maxQ \leftarrow q$
14:         $maxA \leftarrow a$
15:     **end if**
16: **end for**
17: **if** $d = D$ **then**
18:     $bestA \leftarrow maxA$
19: **end if**
20: **return** $maxQ$

---

A recursive approach for tractably estimating $V^*(s, b)$ using a depth-limited online Monte Carlo search is provided in Algorithm 1. Every time the agent needs to execute an action, the function $V(s, b, D, N)$ is called for the current state $s$ and posterior $b$. $D$ corresponds to the depth of the search tree (i.e. planning horizon) and $N$ to the branching factor (i.e. number of successor states to sample at each level, for each action). To sample a successor state $s'$ from $P(s'|s, b, a)$, we can simply sample a graph $G_a$ for action $a$ according to the probabilities $p_a^j$ and then sample $s'$ from this DBN, given that the parents take values $s$. At the fringe, an estimate $\hat{V}(s, b)$ of the return obtained from this posterior is used. Several techniques can be used to estimate $\hat{V}(s, b)$. For instance one could maintain an approximate value function $\hat{V}_j(s)$ for each sampled factored MDP defined by the DBNs $\{(G_a^j, \phi_{G_a^j}) | a \in A\}$ and then compute $\hat{V}(s, b) = \sum_{j=1}^{K} \hat{V}_j(s) \prod_{a \in A} p_a^j$. The approximate value functions $\hat{V}_j(s)$ can be updated efficiently via prioritized sweeping every time the counts $\phi$ are updated. For the experiments presented below, we simply use $\hat{V}(s, b) = \max_{a \in A} R(s, a)$. The UPDATEGRAPHPOSTERIOR updates the Dirichlet posteriors and probabilities $p_a^j$ presuming a transition $(s, a, s')$ was observed. The best action to execute for the current time-step can be retrieved through the $bestA$ variable for the top node of the tree. The computation time allowed to estimate $V^*(s, b)$ can be limited by controlling the branching factor ($N$) and search depth ($D$), albeit at the expense of lesser accuracy.

### 3.3 Resampling DBNs

The current approach is not particularly effective when the initial set of sampled DBN structures is poor, since we are simply updating weights and therefore not changing the structure. This can be addressed by resampling new DBNs from the current posterior $P(G|h)$ to obtain more likely structures after observation of the history $h$. We implement this using an MCMC algorithm, as described in section 2.2. In general, it may not be appropriate to re-sample graphs too frequently. One useful criteria to decide when to resample new graphs is to look at the overall likelihood $L_a$ of our current set of DBNs for a particular action $a$. This can be computed directly from the normalization constant $\eta$ (Equation 8). Presuming that at time $t = 0$, $L_a = 1$, we can simply update $L'_a = \eta L_a$ at every step. Then whenever $L_a$ falls below some threshold, we resample a new set of $K$ graph structures $G_a^{\prime j}$ from posterior $P(G_a|h)$ and update the Dirichlet posterior $P(\theta_{G_a^j} | G_a^j, h)$ for each graph according to the whole history $h$ (starting from the Dirichlet prior $P(\theta_{G_a} | G_a)$). The probabilities $p_a^j$ for these new graphs are then reinitialized to $\frac{1}{K}$ and the likelihood $L_a$ to 1.

## 4 Experiments

To validate our approach, we experiment with instances of the network administration domain (Guestrin et al., 2003). A network is composed of $n$ computers linked together by some topology. Each computer is either in *running* or *failure* mode. A running computer has some probability of transitioning to failure, independent of its neighbors in the network; that probability is increased for every neighbor in failure mode. A computer in failure mode remains so until rebooted by the operator. A reward of +1 is obtained for every running computer in the network at every step, no reward is given for failed computers, and a -1 reward is received for each rebooting action. The goal of the operator is to maximize the number of running computers while minimizing reboots actions. The starting state assumes all computers are running.

In our experiments, we assume a probability $\frac{1}{30}$ that a running computer goes into failed mode and a probability $\frac{1}{10}$ that a failed computer induces failure in any of its neighbors. So at any step, the probability that a running computer remains in a running state is $\frac{29}{30}(0.9)^{N_F}$ where $N_F$ is the number of neighbors in failure state. We assume a discount factor $\gamma = 0.95$.

This problem can be modeled by a factored MDP with $n$ binary state variables, each representing the running state of a computer in the network. There are $n + 1$ actions: a reboot action for each computer and a *DoNothing* action. The DBN structure representing the dynamics when no reboot is performed is a bipartite graph where the state variable $S'_i$ (the next state of computer $i$) depend on $S_i$ (the previous state of computer $i$) and $S_j$ for all computers $j$ connected to $i$ in the network. Note that if $S'_i$ depends on $S_j$, then this implies $j$ is connected to $i$ and thus $S'_j$ depends on $S_i$. Hence the adjacency matrix $\mathcal{A}$ encoding the dependence relations in this bipartite graph, where entry $\mathcal{A}_{ij} = 1$ if $S'_j$ depend on $S_i$, 0 otherwise, is always symmetric and has a main diagonal full of ones.

In terms of prior knowledge, we assume the agent knows that rebooting a computer always puts it back into running mode and doesn't affect any other computer. The goal of the agent is to learn the behavior of each computer in the network when no reboot is performed on them. Therefore, a single DBN is learned for the behavior of the system when no reboot is performed. We also assume the agent knows that the adjacency matrix is symmetric and has a main diagonal of ones. However we do not assume that the agent knows the topology of the network. We choose a prior over structures that is a uniform distribution over bipartite graphs with symmetric adjacency matrix (and main diagonal equal to 1). Given a prior of this form, the set of moves we consider to sample graphs in the Metropolis-Hasting algorithm consist of inverting any of the binary variables in the upper-right half of the adjacency matrix $\mathcal{A}$ (excluding the main diagonal) as well as the corresponding entry in the bottom-left half. Moves of this type preserve the symmetry in the adjacency matrix, and correspond to adding or removing a connection between any pair of computers in the network. We assume no prior knowledge regarding the probabilities of failure, so a uniform Dirichlet prior was used. Under the likelihood equivalence assumption, the prior counts $\phi_G$ are defined such that $\phi_G^{i,v|E} = \frac{1}{|S_{\text{Par}_i(G)}||S_i|} = 2^{-|\text{Par}_i(G)|-1}$.

We consider three different network architectures: a simple linear network of 10 computers (1024 states), a ternary tree network composed of 13 computers (8192 states) and a dense network of 12 computers (4096 states) composed of 2 fully connected components of 6 computers, linked to each other. These networks are shown in Figure 1. To assess the performance of our structured Bayesian RL approach, we compare it to a similar model-based Bayesian RL that learns the full joint distribution table, i.e. the DBN where each next state variable $S'_i$ depends on all previous state variables $S_j$. We also consider the case where the DBN structure is fully known in advance and only the probability parameters are learned. These three approaches are compared in terms of three different metrics:

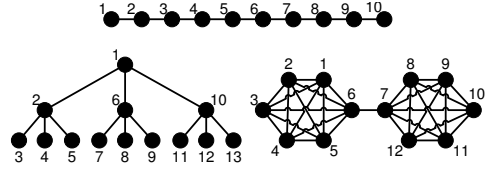

Figure 1: Linear network (top), ternary tree network (left) dense network (right).

empirical return, distribution error and structure error, as a function of the number of learning steps. The distribution error corresponds to a weighted sum of L1-distance between the distributions of the next state variables as defined by the Dirichlet posterior counts and the exact distributions in the system:

$\sum_{j=1}^{K} p_a^j \sum_{s \in S} \sum_{i=1}^{n} \left\| \frac{\phi_{G_a^j}^{i,*|\text{ParVal}_i(s,G_a^j)}}{||\phi_{G_a^j}^{i,*|\text{ParVal}_i(s,G_a^j)}||_1} - P(S'_i|s,a) \right\|_1$.

The structure error is computed as a weighted sum of the errors in the adjacency matrix of the sampled graphs compared to the correct adjacency matrix: $\sum_{j=1}^{K} p_a^j \sum_{i=1}^{n} \sum_{k=1}^{n} |\mathcal{A}_{ik}^{G_a^j} - \mathcal{A}_{ik}^{G^*}|$, where $\mathcal{A}^{G_a^j}$ is the adjacency matrix for sampled graph $G_a^j$ and $\mathcal{A}^{G^*}$ the exact adjacency matrix. All reported results are averaged over 50 simulations of 1500 steps each. Error bars were small, so were removed for clarity.

### 4.1 Linear Network

In the linear network experiment, we sample $K = 10$ graphs, and resampling is performed whenever $\ln L_a < -100$. Online planning is done with depth $D = 2$ and branching factor $N = 5$ for each action. Since we use the immediate reward at the fringe of the search tree, this corresponds to approximate planning over a 3-step horizon. These same parameters are also used for planning with the known structure, and over the full joint probability table. Results are presented in Figures 3-5.

These figures show that our approach (denoted *Structure Learning*) obtains similar returns as when the structure is known in advance (denoted *Known Structure*). Both of these cases reach optimal return (denoted *Known MDP*[1]) very quickly, within 200 steps. Our approach is also able to learn the transition dynamics as fast as when the structure is known a priori. On the other hand, the unstructured approach (denoted *Full Joint*) takes much more time to achieve a good return and learn the dynamics of the system. This confirms that assuming a structured representation of the system can significantly speed up learning. Finally, we also observe that the structure learning al-gorithm is able to learn a good structure of the domain over time (see Figure 4.1). Even though the sampled structures are not perfect, our approach is still able to predict future states of the system with similar accuracy as when the structure is known in advance. The average planning times per action are 100ms for structure learning, and 19ms for the other two approaches with fixed structure.

---
[1]This is the value iteration solution, assuming the structure and parameters are fully known in advance.

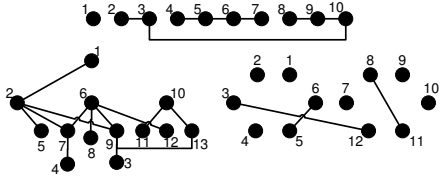

Figure 2: Most likely networks among samples after 1500 steps: Linear network (top), ternary tree network (left) dense network (right).

### 4.2 Ternary Tree Network

In the ternary tree network experiment, we sample $K = 8$ graphs, and resample them whenever $\ln L_a < -150$. For the planning, we use a depth $D = 2$ and sample $N = 4$ next states for each action. Results are presented in Figures 6-8. The results are similar to the Linear Network experiment. The main point to note is that this is a significantly harder problem for the unstructured approach, which even after 1500 steps of learning has not yet improved. This is in contrast to our approach which obtains similar performance as when the structure is known a priori, and reaches optimal performance after just a few hundred steps of learning. These results are obtained even though the priors we provide are very weak. The average planning times per action are 153ms for structure learning, and 29ms for the two approaches with fixed structure.

### 4.3 Dense Network

In the dense network experiment, we sample $K = 8$ graphs, and resample them whenever $\ln L_a < -120$. For the planning, we assume $D = 2$ and $N = 4$. Results are presented in Figures 9-11. In this domain, we observe a surprising result: our approach using structure learning is able to learn the dynamics of the system much faster than when the structure is known in advance (see Figure 10), even though the learned structures are still far from correct (see Figures 11 and 4.1). This is a domain where there are many dependencies between state variables, so there are many parameters to learn (whether or not the structure is known). In such a case, our structure learning approach is at an advantage, because early on in the learning, it can favor simpler structures which approximate the dynam-ics reasonably well from very few learning samples (e.g. $< 250$). As further data is acquired, more complex structures can be inferred (and more parameters estimated), in which case our approach achieves similar return as when the structure is known, while it continues to estimate the true parameters more accurately.

This result has important implications for RL in large domains. Namely, it suggests that even in domains where significant dependencies exist between state variables, or where there is no apparent structure, a structure learning approach can be better than assuming a known (correct) structure, as it will find simple models that allow powerful generalization across similar parameters, thus allowing for better planning with only a small amount of data.

The average planning times per action are 120ms for structure learning, and 22ms for the other two approaches with fixed structure.

## 5 Conclusion

This paper presents a novel Bayesian framework for learning both the structure and parameters of a factored MDP, while also simultaneously optimizing the choice of actions to trade-off between model exploration and exploitation. It is important to note that both the use of a factored representation, and the use of online planning, are key to allowing our approach to scale to large domains. By learning a factored representation, we allow powerful generalization between states sharing similar features, hence learning of the model makes more efficient use of data. It is especially interesting to notice that our structure learning approach is a useful way to accelerate RL even in domains with very weak structure.

### Acknowledgements

This research was supported by the Natural Sciences and Engineering Research Council of Canada (NSERC) and the Fonds Québécois de la Recherche sur la Nature et les Technologies (FQRNT).

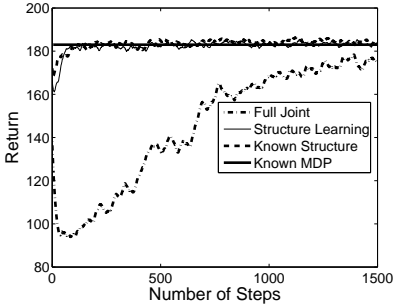
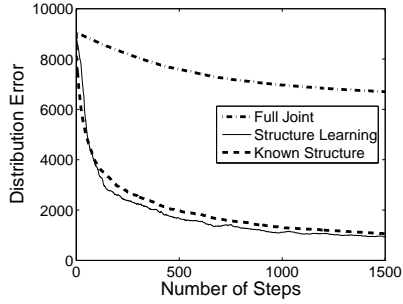
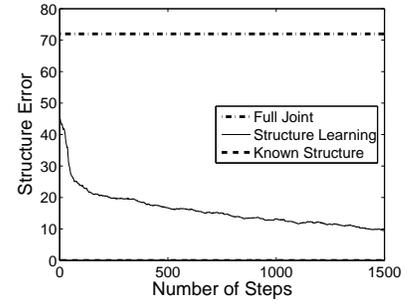

Figure 3: Empirical return in the linear network.

Figure 4: Distribution error in the linear network.

Figure 5: Structure error in the linear network.

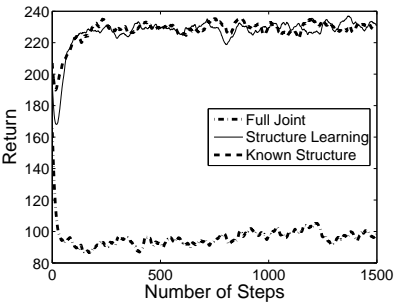
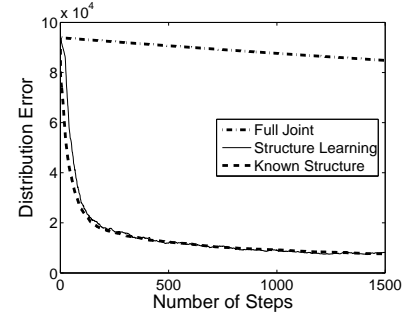
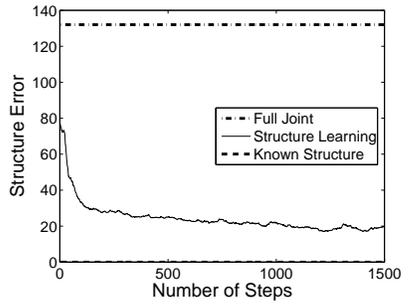

Figure 6: Empirical return in the ternary tree network.

Figure 7: Distribution error in the ternary tree network.

Figure 8: Structure error in the ternary tree network.

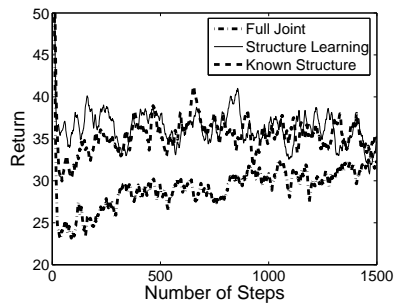
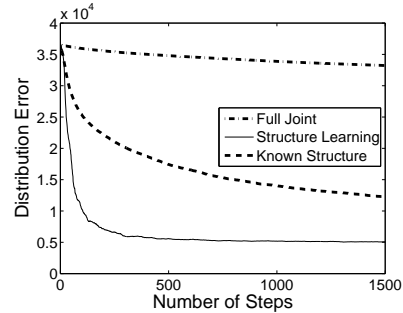
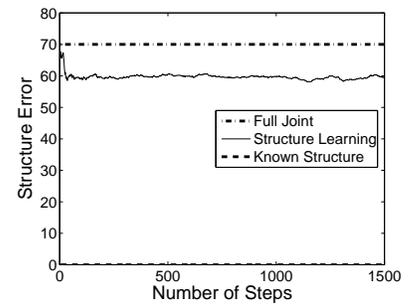

Figure 9: Empirical return in the dense network.

Figure 10: Distribution error in the dense network.

Figure 11: Structure error in the dense network.